\documentclass{article}
\usepackage{arxiv}
\usepackage[utf8]{inputenc}
\usepackage[T1]{fontenc}
\usepackage{hyperref}
\usepackage{url}
\usepackage{booktabs}
\usepackage{amsmath, amssymb, amsfonts}
\usepackage{nicefrac}
\usepackage{microtype}
\usepackage{graphicx}
\usepackage{natbib}
\usepackage{doi}

\usepackage{float}
\usepackage{multirow}   % for multirow headers

\usepackage{subcaption} % for subfigures

\usepackage[linesnumbered,ruled,vlined]{algorithm2e}
\SetKwInOut{KwIn}{Input}
\SetKwInOut{KwOut}{Output}

\usepackage{caption}
\usepackage{siunitx} % for alignment of numbers

\usepackage{tikz}
\usetikzlibrary{positioning, arrows.meta}

\SetAlgoNlRelativeSize{-1}  % smaller line number font
\SetAlgoNlRelativeSize{-1}
\SetAlCapNameFnt{\small}
\SetAlCapFnt{\small}

\title{An Efficient and Almost Optimal Solver for the Joint Routing-Assignment Problem via Path Polishing and Large-$\alpha$ Optimization}

\author{
  {\includegraphics[scale=0.06]{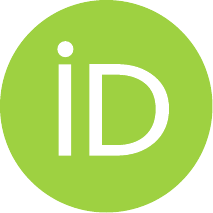}\hspace{1mm} Yuan Qilong } \\
  Singapore Institute of Technology \\  \texttt{qilong.yuan@singaporetech.edu.sg} \\
}

\hypersetup{
  pdftitle={An Efficient and Almost Optimal Solver for the Joint Routing-Assignment Problem via Path Polishing and Large-$\alpha$ Optimization},
  pdfsubject={Optimization, Combinatorics, Gurobi},
  pdfauthor={Yuan Qilong},
  pdfkeywords={Joint Routing–Assignment Optimization, Path Polishing, Large-$\alpha$   Constraints, Mixed Integer Programming},
}

\begin{document}
\maketitle

\begin{abstract}
The \textit{Joint Routing-Assignment (JRA)} optimization problem simultaneously determines the assignment of items to placeholders and a Hamiltonian cycle that visits each node pair exactly once, with the objective of minimizing total travel cost. Previous studies introduced an exact mixed-integer programming (MIP) solver, along with datasets and a Gurobi implementation, showing that while the exact approach guarantees optimality, it becomes computationally inefficient for large-scale instances. To overcome this limitation, heuristic methods based on merging algorithms and shaking procedures were proposed, achieving solutions within approximately 1\% deviation from the optimum.
This work presents a novel and more efficient approach that attains high-accuracy, near-optimal solutions for large-scale JRA problems. The proposed method introduces a \textit{Partial Path Reconstructon (PPR)} solver that first identifies key item-placeholder pairs to form a reduced subproblem, which is solved efficiently to refine the global solution. Using this PJAR framework, the initial heuristic merging solutions can be further improved, reducing the deviation by half. Moreover, the solution can be iteratively polished with PPR based solver along the optimization path to yield highly accurate tours. Additionally, a global \textit{Large-$\alpha$} constraint is incorporated into the JRA model to further enhance solution optimality. Experimental evaluations on benchmark datasets with $n = 300$, $500$, and $1000$ demonstrate that the proposed method consistently delivers almost optimal solutions, achieving an average deviation of $0.00\%$ from the ground truth while maintaining high computational efficiency. Beyond the JRA problem, the proposed framework and methodologies exhibit strong potential for broader applications. The \textit{PPR} method and its variations as while as \textit{Large-$\alpha$} method can be naturally extended to the classical Traveling Salesman Problem (TSP) and related routing and logistics optimization problems. YouTube Video: \url{https://www.youtube.com/watch?v=2hbskZhwO-k&list=PLpiFbst-QlpjtRRbjB5-lJl4AN778Ic-5}
\end{abstract}

\keywords{Joint Routing–Assignment Almost Optimal Solver \and Path Polishing \and Large-$\alpha$ - Constraints \and Mixed Integer Programming}

\section{Introduction}
\label{sec:intro}

Many practical tasks in robotics, logistics, and manufacturing require transporting a collection of discrete objects from their current locations to designated target places. 
Such problems naturally couple three interdependent decision layers: 
(i) \emph{assignment} - determining which object should go to which labeled placeholder; 
(ii) \emph{sequencing} - deciding in what order the agent should pick up and deliver the objects; and 
(iii) \emph{routing or motion planning} - finding a feasible path that minimizes travel cost while satisfying geometric and kinematic. This paper focus on first two concerns.

Solving the Joint Routing-Assignment (JRA) problem is challenging but has significant practical value. Tabletop manipulation tasks can be highly complex. For example, preparing a table for a meal involves selecting dozens of items and placing them in specific locations. How can such tasks be performed efficiently by a robotic system? In these scenarios, items may not be unique, and additional constraints, such as placement compatibility or sequence requirements, must be considered to ensure feasible solutions. Even more sophisticated scenarios exist in board-game-like tasks; for example, in the game of Go, more than 300 pieces may need to be arranged. Developing a JRA solver provides a solid foundation for addressing more complex problems in robot task planning and execution.  

In practical item packaging, collection, or arrangement tasks, a large number of objects must be picked in a specific order and placed in designated locations according to their type. For tabletop robotic manipulation systems, finding an optimal path is crucial, as minimizing the total path length directly reduces time and energy consumption. Typically, the robot arm picks an item, places it in the corresponding placeholder, and repeats this process until all items are allocated, after which it returns to its rest position. Efficiently solving such routing and assignment problems is therefore essential for high-performance robotic operation.

This problem is closely related to the classical Traveling Salesman Problem (TSP), but it is significantly more complex due to the combinatorial nature of item-to-location assignments. Many practical applications naturally fall into this model. Examples include:

\begin{enumerate}
    \item \textbf{Item packaging:} Allocating unique items (e.g., mechanical parts, fruits, cakes, toys, or other commercial products) to containers, where each item type must be placed in containers with multiple placeholders.
    \item \textbf{Screw or pin assembly:} Sequentially picking screws or pins and assembling them into target holes, with no restrictions on which item must go to which hole.
    \item \textbf{Robot planting:} Continuously picking plants and placing them into designated planting locations.
    \item \textbf{Room tidying:} Picking up items and placing them in feasible locations, where many items must be handled in a sequential manner.
\end{enumerate}

Although compactly stated, this problem integrates the combinatorial structure of assignment and routing with precedence and connectivity constraints. The decision variables typically include a bijective mapping between items and placeholders, as well as a permutation representing the robot’s tour. The robot must respect unit-capacity and pickup-before-delivery constraints. The objective is generally to minimize total travel distance, which reflects both time and energy consumption.

Efficiently handling such tasks with a robotic system requires accounting for non-unique items and additional constraints to ensure feasible solutions. Developing accurate and efficient JRA solvers provides a baseline framework for addressing these problems and serves as a foundation for creating solvers for more complex variations and sophisticated application scenarios. This work can help users design robotic systems that achieve efficient task planning and execution.

\subsection{Connections to existing problem classes}
This problem lies at the intersection of several established families in operations research and robotics.

\paragraph{Pickup-and-delivery and the stacker-crane problem.}
Pickup-and-delivery variants of the traveling salesman and vehicle routing problems (PDTSP, VRP-PD) enforce precedence constraints between pickup and delivery pairs under vehicle capacity limits. 
The stacker-crane abstraction \cite{hernandez2004,hernandez2007,treleaven2012} is closely related, representing a single-vehicle system that must transport items between paired pickup and drop-off locations. 
These problems are NP-hard in general and have motivated a rich literature on branch-and-cut formulations, exact algorithms for small-scale instances, and heuristic or metaheuristic methods for larger ones. 

\paragraph{Assignment - routing integration.}
Classical formulations of the assignment problem \cite{kuhn1955hungarian} and the traveling salesman problem \cite{dantzig1954solution, lawler1985traveling, applegate2006traveling} address individual subcomponents of our task. 
However, when assignment and routing are coupled the search space expands combinatorially, requiring integrated formulations or hierarchical decomposition. 
Recent robotics literature has explored mixed-integer and sampling-based approaches for such coupled problems, yet scalability and global optimality remain open challenges. 
Modern solvers such as Gurobi \cite{gurobi2024manual} allow constraint families to be activated dynamically through cutting-plane method based callbacks, a strategy that has been successfully employed in subtour elimination for TSPs and connectivity-enforced routing \cite{applegate2006traveling}. 
This mechanism reduces model size and computational overhead by adding connectivity constraints only when violated, which we exploit in our implementation. %However, results show that such methods is still not efficient enough to couple with large size problems when the number of pairs is more than 300.

\subsection{Purposes of the Article}

In previous work, the authors introduced an exact MIP based solver and shared the datasets together with Gurobi implementation code and experimental results showing that the solver is lack of efficiency in solving large-size JRA problem \cite{yuan2025datasets}. Subsequently, heuristic solvers \cite{qilong2025assignmentroutingoptimizationefficient} based on heuristic merging algorithms and shaking method have been introduced which provide solution within 1 percent deviation error from ground truth reference. 

This work is an approach to better resolve the problem with high accuracy and high efficiency such that it can provide almost optimal solution efficiently for large-size JAR problem. 

Existing algorithms such as 2-opt, 3-opt, k-opt and Lin-Kernighan-Helsgaun  (LKH) algorithms \cite{Helsgaun2009GeneralKS} improve the path through breaking connections and reconnect through less cost approaches, normally the number of edges to breaking is less than 10 because of k-opt algorithm with $k$ edges exchanges, $(k-1)!*2^{(k-1)}$ reconnection approaches exists in the possible options. However, based on experiments as mentioned in Section 5, it is noted that path improvements can involve much larger edge-changes. In other words, a k-opt optimal solution is not always able to progress towards optimal in path improvements. Such facts motivate us to work on new methods to resolve such problem with new approaches for optimal or at least almost optimal solutions.

Inspired by the cycle merging algorithm as introduced in \cite{qilong2025assignmentroutingoptimizationefficient}, it is noted that path improvement happens with edges breaking and reconnection at different regions at the path domain, which is closely related to the geometry of the tour path. Therefore the authors are inspired to solving the problem through applying circle filters along the path to select nodes within the circle. Subsequently, the path is further optimized based on breaking the connections within/ around the circle, and find optimal way for  reconnection based on a newly introduced efficient JAR solver. Such solver can
iteratively improve the solution along the path. Together with introduced initial guess solution estimation and final optimization methods, we achieve almost optimal solution for large size (upto 1000 pairs have been tested) within comparable computation time
as compared with heuristic methods while providing much more accurate results (0.00\% deviation error from ground truth). 

\begin{figure}[h]
  \centering  \includegraphics[width=0.8\linewidth]{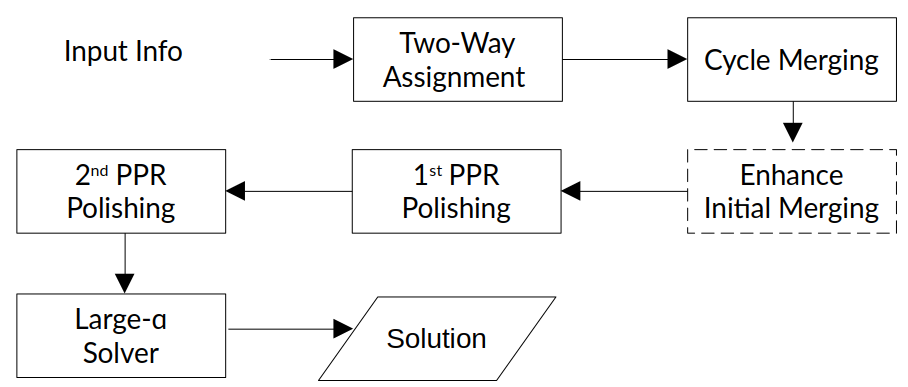}  
  \caption{Illustration of Workflow of The Proposed Hungarian algorithm}
  \label{fig:flow}
  \vspace{0mm}
\end{figure}
The overall method is illustrated in Figure \ref{fig:flow}, initial guess solution is estimated from two-way assignment method based on Hungarian algorithm and cycle merging algorithm as introduced in \cite{qilong2025assignmentroutingoptimizationefficient}. 
With a newly proposed partial JAR solver which firstly select the key item-placeholder pairs and form a smaller size partial JAR (PJAR, actual PPR as introduced in Section 2.2) problem to efficient update the solution, initial guess solutions can be improved. The introduced following processes are in general as follows:

\begin{enumerate}
    \item Enhancing the initial merging with PJAR Base on such PJAR method. 
    \item The solution is polished along the path  with circular filters to select nodes for edge breaking and reconnection using such PJAR solver to achieve very high accuracy path. 
    \item Finally, based on the high quality path, a \textbf{Large}-$\alpha$ constrain is added to JAR model to further improve the solution efficiently to almost optimal solution with 0.00 percent deviation error from ground truth.
\end{enumerate}

The paper is organized in following manner. Section 2 introduces the problem formulation and the methodology in problem solving. Section 3 introduces the computational complexity of methods. Section 4 presents the experimental setup and the results. Section 6 concludes the paper.

\section{Mathematical Formulation}
For the Joint Assignment-Routing Problem with $n$ items and $n$ placeholders, let $I = \{1, \dots, n\}$ be the set of items, $P = \{n+1, \dots, 2n\}$ the set of placeholders, and $c_{ij}$ the cost of traversing edge $(i,j)$. 

Denote $V = I \cup P$ and $E = \{(i,j) \in V \times V : i \neq j\}$. 

We define binary decision variables:
\[
x_{ij} =
\begin{cases}
1, & \text{if edge } (i,j) \text{ is included in the cycle}, \\[4pt]
0, & \text{otherwise.}
\end{cases}
\]

The objective is to minimize total traversal cost :
\[
\min \sum_{(i,j) \in E} c_{ij} x_{ij},
\]
while the solution form a large cycle and the path ensures that every items and placeholders take turns to show up along the path, which are discussed in next subsection. 
\subsection{Simplified Model Constraints}

In order to solve the problem efficiently, in this simplified modeling, only edges from item to placeholder is allowed.  Constraints are as follows:
\begin{enumerate}
    \item \textbf{Non-connectable constraints edges:} The constraints defined for non-connectable pairs are as follows:
    \[
    x_{ij} = 0 \text{ if both } i,j \in I \text{ or } i,j \in P \text{ or } ( i \in P  \text{ and } j \in I)
    \]
    % \[
    % \sum_{j \in V, j \neq i} x_{ij} + \sum_{j \in V, j \neq i} x_{ji} = 2, \quad \forall i \in V
    % \]
    \item \textbf{Assignment:} Each item and placeholder is assigned twice:
    \[
    \sum_{p \in P} x_{ip} = 2, \quad \forall i \in I, 
    \quad
    \sum_{i \in I} x_{ip} = 2, \quad \forall p \in P
    \]
    \item \textbf{Fixed pair:} In case of having starting and goal points, we specify starting position to be the last placeholder. The goal point is specified to be the last item and is enforced to be assigned to the last placeholder as following constraints. 

\[
x_{n, 2n} = 1
\]
    \item \textbf{Subtour elimination:} Disconnected cycles are prohibited by a cutting-plane method \cite{kelley1960cutting,westerlund1995extended}:
    \[
    \sum_{i,j \in S, i \neq j} x_{ij} \le |S|-1, \quad \forall \text{ subtours } S \subset V
    \]
\end{enumerate}
This is enforced dynamically using subtour elimination callback function in Gurobi.
\subsection{Path Improvement: Partial Path Reconstruction (PPR) Method}
As discused in \cite{gurobi2025mipprimer}, exact solver implementation has low efficiency for large-size JAR problem. If we can improve heuristic solutions efficiently and largely enhance the accuracy, it would be very meaningful for JAR problem solving. Therefore, the idea here is to get a initial guess solution first and then propose method to efficiently identify possible areas to break the connections and further optimize the solution with new connection solutions. 

Given an initial feasible large cycle (tour) defined over the sets of items and placeholders, the current solution can be represented by two ordered lists:
\[
q_I = [i_1, i_2, \dots, i_n], \quad 
q_P = [p_1, p_2, \dots, p_n],
\]
where $q_I(k) \in I$ denotes the $k$-th item and $q_P(k) \in P$ denotes the $k$-th placeholder in the sequence.

Each item $q_I(k)$ is connected to its two adjacent placeholders, namely $q_P(k-1)$ and $q_P(k)$, thus:

\begin{equation}
x_{q_I(k), q_P(k-1)} = 1, \quad 
x_{q_I(k), q_P(k)} = 1,
\quad \forall k = 1, \dots, n.
\label{eq:initial-cycle-edges}
\end{equation}

where indices are taken modulo $n$ for the cyclic structure.

\subsubsection{Breaking and Decomposition of the Cycle.}
Starting from a initial guess, if we identify a certain domain and break the edges within/around it, there are possibly more optimal reconnection approaches. As such domains is geometrically dealing with clusters of neighborhood nodes which can be more than 10 pairs, such method can potentially improve the path outperforming k-opt when the optimization solver is nicely designed. 

\textbf{Breaking:} If a subset of nodes $V_S \subseteq V = I \cup P$ is selected for disconnection, all edges incident to these nodes are removed:
\[
x_{ij} = 0, \quad \forall (i,j) \in E : i \in V_S \text{ or } j \in V_S.
\]
The removal of these edges partitions the original tour into several \emph{partial paths}, each representing a continuous linkage of alternating items and placeholders.  
Let the resulting set of partial paths be denoted by
\[
\mathcal{Q} = \{Q_1, Q_2, \dots, Q_m\},
\]
where each $Q_s = (q_{I_S}, q_{P_S})$ corresponds to one such linkage.

\paragraph{Identification of Segments.}
For each segment $Q_s$, we distinguish:
\begin{enumerate}
    \item The ordered list of items $q_{I_S} = [i_{s,1}, i_{s,2}, \dots, i_{s,k_s}]$
    \item The corresponding placeholders $q_{P_S} = [p_{s,1}, p_{s,2}, \dots, p_{s,k_s}]$ that form the linkage.
\end{enumerate}

To facilitate subsequent model updates and local improvement, only the \emph{boundary nodes} of each segment are retained for reconnection:
\[
i_s^{\text{start}} = i_{s,1}, \quad p_s^{\text{end}} = p_{s,k_s}.
\]
These boundary nodes define the admissible reconnection points for future optimization iterations.
The boundary pairs and the separated nodes form the new set of items and placeholders for optimization. \textbf{Here, in case the stopping points and starting point are within the selections, linking constraints as introduced in Section 2.1 are maintained.} 
\paragraph{Reserved Edge Set.}
To ensure solution recoverability and preserve valid linkages already present in the partial paths, all edges belonging to the internal structure of each $Q_s$ are stored in a reserved edge list:
\[
L_r = \{ (i,j) \in E : x_{ij} = 1 \text{ and } (i,j) \in Q_s, \ \forall Q_s \in \mathcal{Q} \}.
\]
The reserved edge set $L_r$ is fixed during subsequent optimization runs and reintroduced into the model to maintain the integrity of existing substructures:
\[
x_{ij} = 1, \quad \forall (i,j) \in L_r.
\]
This mechanism allows for local refinement and efficient re-optimization without losing valuable structural information from the previous feasible tour.

\paragraph{Summary.}
The proposed update process can be summarized as follows:
\begin{enumerate}
    \item Start from an initial tour defined by $(q_I, q_P)$.
    \item Select a subset of nodes $V_S$ for disconnection.
    \item Remove all incident edges to $V_S$ to create partial paths $\mathcal{Q}$.
    \item Record boundary nodes $(i_s^{\text{start}}, p_s^{\text{end}})$ for each segment.
    \item Store all existing internal edges in $L_r$ for later recovery.
\end{enumerate}
This approach allows the model to be dynamically updated, focusing computational effort on unconnected or modified regions, while maintaining the consistency of existing valid linkages.

\subsubsection{Solution Recovery and Reconstruction of the Complete Tour}
Let function $P_{solver}$ denote the Partial Path Reconstruction solver. Thus, we have, 
\begin{equation}
\label{eq:PPRSolver}
L_n = P_{solver}(\mathcal{Q},L_r,V_S )
\end{equation}

which means that performing the partial optimization described above, the solver returns an updated set of active edges
\[
L_n = \{ (i,j) \in E : x_{ij} = 1 \text{ in the partial solution} \}.
\]
This set represents the optimized edges obtained from the reduced problem that only considered reconnections among boundary nodes and unlinked points.

TODO: Add Figure descriptions

\paragraph{Removal of Temporary Linkages.}
During the partial optimization process, temporary edges are introduced to connect the starting item and ending placeholder of each previously existing linkage.  
These artificial edges ensure that the reduced problem remains connected (with each boundary nodes maintain one more degree to be connected for the new solver) but do not correspond to valid traversals in the final full-cycle solution. Therefore, such temporary connection need to be removed during entire path reconstruction after path reconnection optimization process. 

$L_t$ denote the set of these temporary edges.  
For each linkage segment $Q_s = (q_{I_S}, q_{P_S})$:
\[
L_t \subseteq \{ (i_s^{\text{start}}, p_s^{\text{end}}) \},
\]
where $(i_s^{\text{start}}, p_s^{\text{end}})$ is the auxiliary edge linking the beginning item and ending placeholder of the segment.

However, in the case where a segment $Q_s$ contains only a single pair $(i,p)$, i.e.,
\[
|q_{I_S}| = |q_{P_S}| = 1,
\]
no such temporary edge exists, and therefore no removal is needed.

\paragraph{Reintegration of Reserved Linkages.}
The reserved edges $L_r$  representing the preserved valid connections which remain unchanged by the partial optimization.  
These edges are reintroduced into the solution to reconstruct the complete tour. Therefore, the final set of active edges that defines the complete tour is thus obtained by:
\[
L^{*} = (L_n \setminus L_t) \cup L_r.
\]
The corresponding binary decision variables can be recovered as:
\[
x_{ij}^{*} =
\begin{cases}
1, & \text{if } (i,j) \in L^{*},\\[3pt]
0, & \text{otherwise.}
\end{cases}
\]

\paragraph{Cycle Formation and Post-processing.}
The reconstructed edge set $L^{*}$ is then passed to a cycle reconstruction function, denoted as:
\begin{equation}
\label{eq:c2ip}
q_I , q_P= \text{CyclePath}(L^{*}),
\end{equation}

which returns the ordered item - placeholder sequence forming a valid (Hamiltonian) big cycle (or complete tour).

\paragraph{Summary.}
The complete solution recovery process is summarized as:
\begin{enumerate}
    \item Obtain partial solution edge set $L_n$ from the solver.
    \item Identify and remove temporary linkage edges $L_t$.
    \item Form the unified set $L^{*} = (L_n \setminus L_t) \cup L_r$ to update the edges for final path.
    \item Apply the reconstruction function $\text{CyclePath}(L^{*})$ to obtain the full tour.
\end{enumerate}

This reconstruction procedure guarantees that the final solution preserves all previously reserved linkages, integrates improved connections from the partial optimization, and yields a coherent single-cycle tour consistent with the original problem structure.
 
\subsection{Initial Merging Solution}
As introduced in \cite{qilong2025assignmentroutingoptimizationefficient}, it is very fast to obtain an initial merging solution based on the following two steps:
\begin{enumerate}
    \item Apply Hungarian algorithm to assign item to placeholders to get the minimal sum of distance cost assignment. 
    \item Add constraints and apply back assignment from placeholder to items. 
    \item Apply a cycle detector to collect all the cycles based on two-step assignment results.
    \item Iteratively merge the cycles based on least distance cost incremental approach, until only one big cycle is remained.
\end{enumerate}

\begin{figure}[h]
    \centering
    
    \begin{subfigure}[b]{0.48\textwidth}
        \centering
        \includegraphics[width=\textwidth]{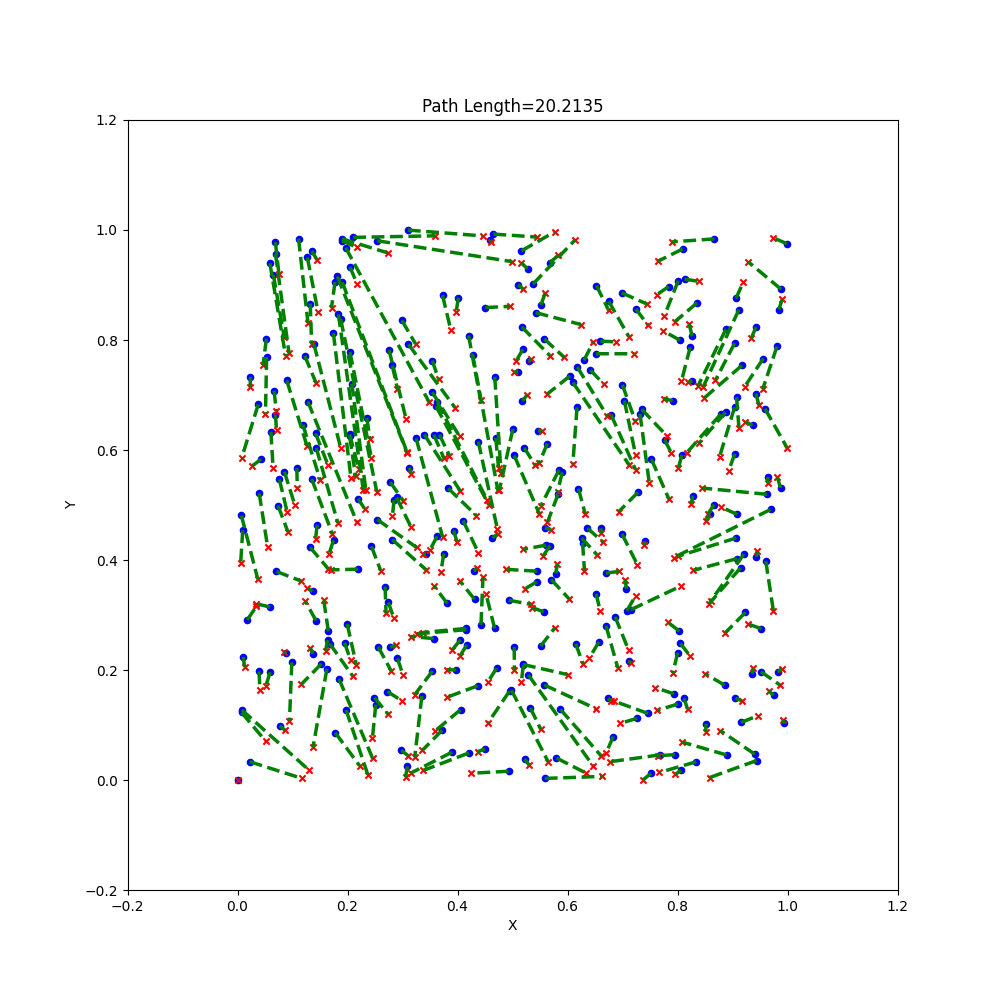}
        \caption{Assign Items to Placeholders}
        \label{fig:one}
    \end{subfigure}
    \begin{subfigure}[b]{0.48\textwidth}
        \centering
        \includegraphics[width=\textwidth]{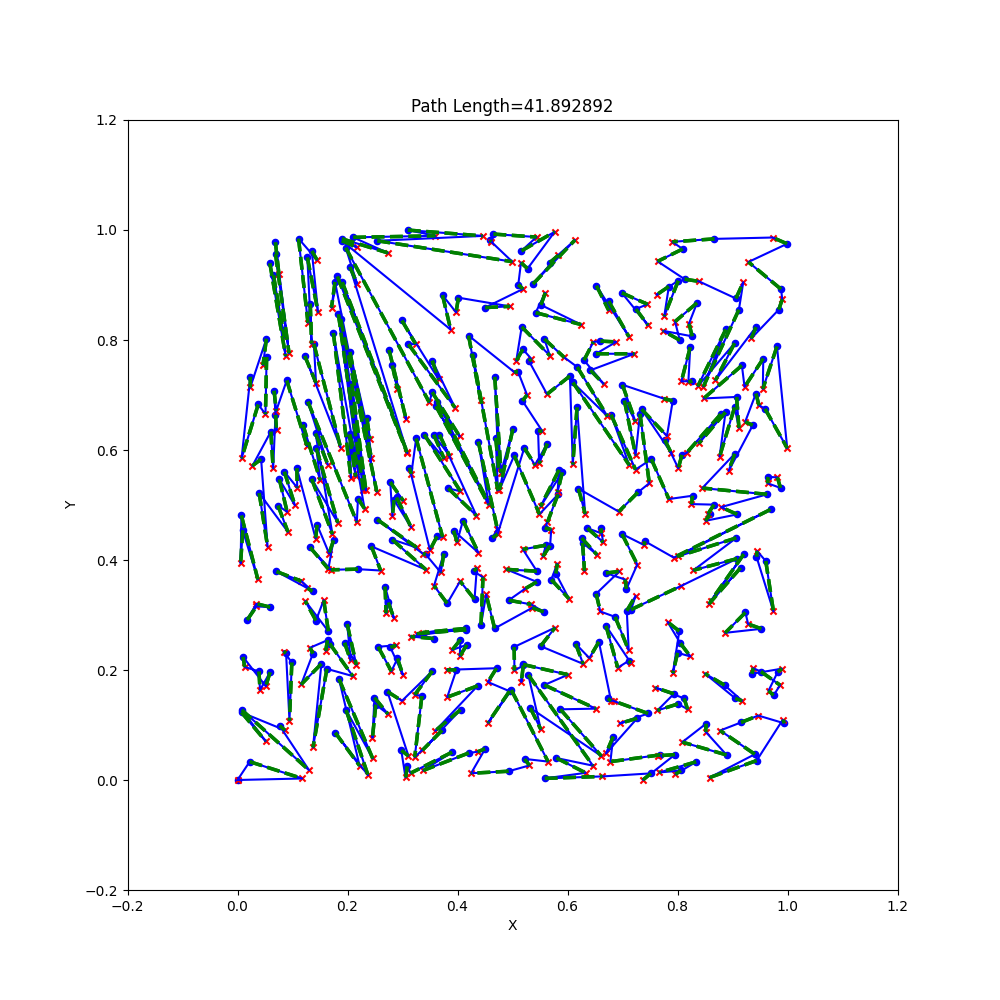}
        \caption{ Assignment from Placeholder Back to Items}
        \label{fig:two}
    \end{subfigure}
    \begin{subfigure}[b]{0.48\textwidth}
        \centering
        \includegraphics[width=\textwidth]{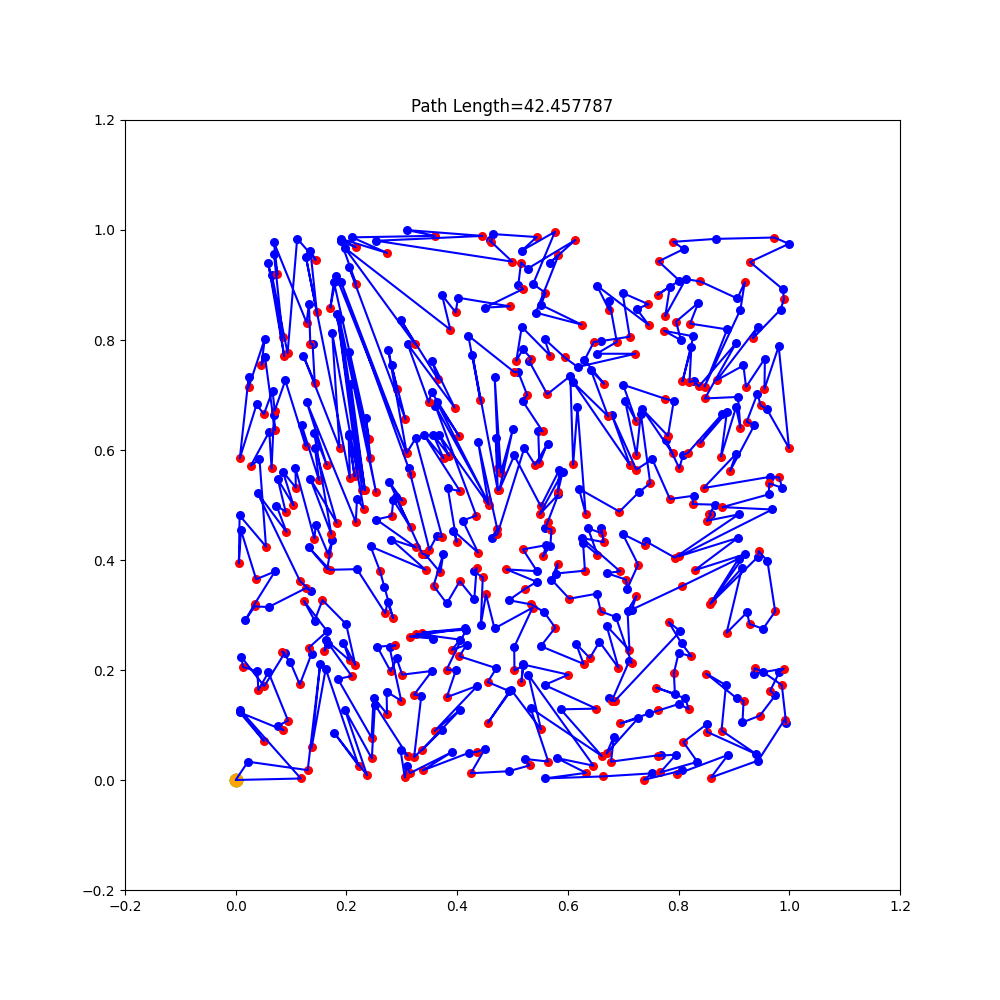}
        \caption{Cycle Merging based Initial Guess}
        \label{fig:three}
    \end{subfigure}
    
    \caption{Three figures displayed side by side.}
    \label{fig:threefigs}
\end{figure}
Figure shows an example of initial merging results for $n=300$ sample. Based on study in \cite{qilong2025assignmentroutingoptimizationefficient}, it is noted that the accuracy of final solution from the heuristic algorithm is restricted by such initial guess especially when the problem size is very large and follow-up processing methods have difficulty in globally improving the solution. This motivates us to: 1. work on new merging method: to apply the introduced Partial Path Reconstruction (PPR) method to get high quality merging results. 2. research on methods for better path improvement algorithm from an initial guess that is fast to calculate. 

\begin{algorithm}[H]
\label{algo:merge}
\caption{Simplified Cycle Merging with Node Collection}
\KwIn{Set of cycles $\mathcal{C} = \{C_1, C_2, \dots, C_m\}$}
\KwOut{Merged cycle $C$, node collector $N_{\mathrm{clc}}$}

Initialize $N_{\mathrm{clc}} \gets \emptyset$\;
\While{$|\mathcal{C}| > 1$}{
  Figure out the two cycles $C_a, C_b \in \mathcal{C}$ to merge with list cost incremental\;
  Identify nodes $(i_a, j_a)$ in $C_a$ and $(i_b, j_b)$ in $C_b$\;
  Form $N_{\text{tmp}} = \{i_a, j_a, i_b, j_b\}$\;
  Update $N_{\mathrm{clc}} \gets N_{\mathrm{clc}} \cup N_{\text{tmp}}$\;
  Remove edges $(i_a, j_a)$ and $(i_b, j_b)$; add $(i_a, j_b)$ and $(i_b, j_a)$\;
  Merge cycles and update $\mathcal{C}$\;
}
Set $C$ as the final remaining cycle\;
\Return $(C, N_{\mathrm{clc}})$\;
\end{algorithm}

The \textbf{Cycle Merging with Node Collection} algorithm is applied after the cycle detection stage, where multiple disjoint cycles $\mathcal{C} = \{C_1, C_2, \dots, C_m\}$ have not been merged to a directed complete tour. 
Each cycle $C_k$ represents a locally valid sub-tour obtained from the forward and backward assignment steps.

Referring to \cite{qilong2025assignmentroutingoptimizationefficient}, the cycle merging and node collection algorithm is shown in algorithms \ref{algo:merge}. The algorithm iteratively merges these sub-cycles until a single large cycle is formed, ensuring that all items and placeholders are connected into one coherent traversal. 
At each merge step, the four connecting nodes involved in the reconnection are collected into the node collector set $N_{\mathrm{clc}}$, which records all affected nodes throughout the merging process.

The outputs of this cycle merging procedure are:
\begin{itemize}
    \item The final merged large cycle $C$, representing a complete feasible tour over $V = I \cup P$;
    \item The node collector $N_{\mathrm{clc}}$, containing all nodes that participated in any merge operation. 
\end{itemize}

Thus, this algorithm not only produces the unified tour but also provides node set information essential for iterative improvement of feasible solutions.

\subsection{Tour Optimization: Partial Path Reconstruction Method}

After obtaining the merged global cycle $C$ and the corresponding node collector 
$N_{\mathrm{clc}}$ from the cycle merging procedure, a pre-processing function 
denoted by $G_{\mathrm{PPR}}$ is applied to transform the current complete tour 
into a new optimization instance of the \textit{Partial Path Reconstruction (PPR)} problem.

\paragraph{Identification of Break Points.}
Nodes contained in $N_{\mathrm{clc}}$ are treated as potential break points where the current tour can be decomposed into smaller subpaths for local reconstruction.  
For each node $v \in N_{\mathrm{clc}}$, its incident edges in the cycle are temporarily removed:
\[
x_{v, u_1} = x_{v, u_2} = 0, \quad \forall (v,u_1), (v,u_2) \in C.
\]
As a result, the full cycle $C$ is partitioned into a set of \textit{partial paths}:
\[
\mathcal{Q}_{\mathrm{PPR}} = \{ Q_1, Q_2, \dots, Q_k \},
\]
where each $Q_s$ denotes a contiguous subsequence of alternating items and placeholders. Proper allocation of placeholders at the boundary are needed.  

Each partial path $Q_s$ can be described by the ordered pair of its boundary nodes:
\[
Q_s = \big( i_s^{\mathrm{start}}, p_s^{\mathrm{end}} \big),
\]
with $i_s^{\mathrm{start}} \in I$ denoting the first item node and 
$p_s^{\mathrm{end}} \in P$ denoting the last placeholder node of the segment.  
All internal edges within $Q_s$ remain fixed and are preserved in the reserved edge set:
\[
L_r = \{ (i,j) \in E : x_{ij} = 1 \text{ and } (i,j) \in Q_s, \ \forall Q_s \in \mathcal{Q}_{\mathrm{PPR}} \}.
\]

\paragraph{Output and New Optimization Domain.}
The preprocessing function $G_{\mathrm{PPR}}$ outputs the data structures required to define a new Partial Path Reconstruction problem:
\[
G_{\mathrm{PPR}}(C, N_{\mathrm{clc}}) \rightarrow \big( \mathcal{Q}_{\mathrm{PPR}}, L_r, N_{\mathrm{clc}} \big),
\]
which together form the reduced problem domain for re-optimization as introduced in Section 2.2.  
Specifically:
\begin{itemize}
    \item $\mathcal{Q}_{\mathrm{PPR}}$ provides the partial path segments to be reconnected;
    \item $N_{\mathrm{clc}}$ specifies the selected nodes eligible for reconnection.
    \item $L_r$ contains all reserved internal edges that must remain fixed;
\end{itemize}

As a matter of fact, sometimes merging multiple cycles together is more efficient than merging one pair at a time. Therefore, by restricting re-optimization to the boundary nodes $N_{\mathrm{clc}}$ while preserving internal path structures $L_r$, computational effort is greatly reduced without compromising overall solution feasibility. 

In this method, the introduced Gurobi JAR solver is applied as the $P_{solver}$ function for reconnection optimization.  As we notice in Table \ref{tab:cycle_percentage_combined} that the cycle size for problem with size $n$ have about $n/5$, and the merging problem result to be of size $1/3$ of the complete node size, which is significantly lighter than original problem and thus much more efficient to solve. 

\subsection{Spatially Localized Partial Path Reconstruction (SLPPR)}

To enable targeted improvement of the existing tour, a spatially localized refinement mechanism is introduced. 
Let $P_g = (x_g, y_g)$ denote a selected geometric point in the planar domain, and let $r_r > 0$ represent the refinement radius that defines a circular region:
\[
\Omega_r = \{ (x, y) \in \mathbb{R}^2 : \| (x, y) - P_g \|_2 \le r_r \}.
\]

\paragraph{Selection of Nodes.}
Let each node $v \in V = I \cup P$ have spatial coordinates $(x_v, y_v)$.  
All nodes located within the refinement region $\Omega_r$ are selected as:
\[
V_r = \{ v \in V : \| (x_v, y_v) - P_g \|_2 \le r_r \}.
\]
These nodes represent the local subset of items and placeholders that are subject to reconnection or reassignment. Here again, the remaining linkage segments are processed to record of edges to be maintained and at the same time generate the temporary boundary item-placeholder pairs to be used for further path optimization through PPR as introduced in Section 2.2. At boundary distal node, it can be an item node or placeholder one. Linking constraints can be added in case both distal are of the same type during the MIP module construction. 

Subsequently, the local path improvement can be achieved about the specific circular domain based on such Local-PPR method. 
Since the solution tour is a complete path, here the authors propose to iteratively polish the solution with such Local-PPR
along the path selecting way-points every other few steps $N_{stp}$.

\subsubsection{Estimation of the Refinement Radius}

The radius $r_r$ serves as the key parameter defining the computational size of each Local-PPR problem.  
To maintain a manageable problem scale, the radius can be set based on experimental fine-tuning, or can be determined adaptively such that the expected number of selected nodes within the refinement region is around $n_{i_n}$.

\paragraph{Expected Node Density.}
Let the total number of nodes in the problem be $n$, and let the overall domain of interest have an effective area $S$.  
The average spatial density of nodes can then be approximated as: $\rho = \frac{n}{S}$.

The expected number of item-nodes (size will be doubled when placeholder are included) located within this circular region can therefore be expressed as
\begin{equation}
n_{i_n} \approx \rho \, S_r = \frac{n}{S} \, \pi r_r^2.
\label{eq:node-estimation}
\end{equation}

\paragraph{Radius Determination.}
Rearranging Equation~\eqref{eq:node-estimation} yields the following expression for the radius $r_r$ in terms of the target node count $n_{i_n}$:
\begin{equation}
r_r = \sqrt{\frac{n_{i_n} \, S}{\pi n}}.
\label{eq:radius-general}
\end{equation}
In practice, this radius provides an adaptive mechanism to control the local problem size.  
For example, if the spatial scale of the domain $S$ is normalized (e.g., $S = 1$), the radius simplifies to
\begin{equation}
r_r = \sigma \, \frac{1}{\sqrt{n}},
\label{eq:radius-simple}
\end{equation}
where $\sigma$ is a proportionality constant that can be tuned to adjust the effective size of the Local-PPR subproblem.

Equation~\eqref{eq:radius-general} shows that the refinement radius grows with the square root of the target number of nodes $n_{i_n}$ and inversely with the square root of the global node density.  
Therefore:
\begin{itemize}
    \item A larger radius $r_r$ corresponds to a larger local optimization region, potentially improving solution quality but increasing computational cost.
    \item A smaller radius $r_r$ focuses refinement on tighter local neighborhoods, yielding faster but more localized adjustments.
\end{itemize}

This adaptive relationship between $r_r$ and $n_{i_n}$ ensures a balanced trade-off between solution quality and computational efficiency, allowing the Local-PPR framework to scale effectively with the problem size.

\paragraph{Summary.}
The spatially localized PPR process can thus be summarized as follows:
\begin{enumerate}
    \item Specify a location $P_g$ and radius $r_r$ to define refinement region $\Omega_r$;
    \item Select nodes $V_r$ within $\Omega_r$ and remove incident edges;
    \item Identify unselected linkage segments $C_{\mathrm{u}}$ and store their internal edges in $L_r$;
    \item Form and solve a local PPR problem restricted to $V_r$ and $C_u$;
    \item Integrate the updated local solution into the global tour and recover the new full path.
\end{enumerate}

\begin{figure}[h!]
\centering    \includegraphics[width=0.8\textwidth]{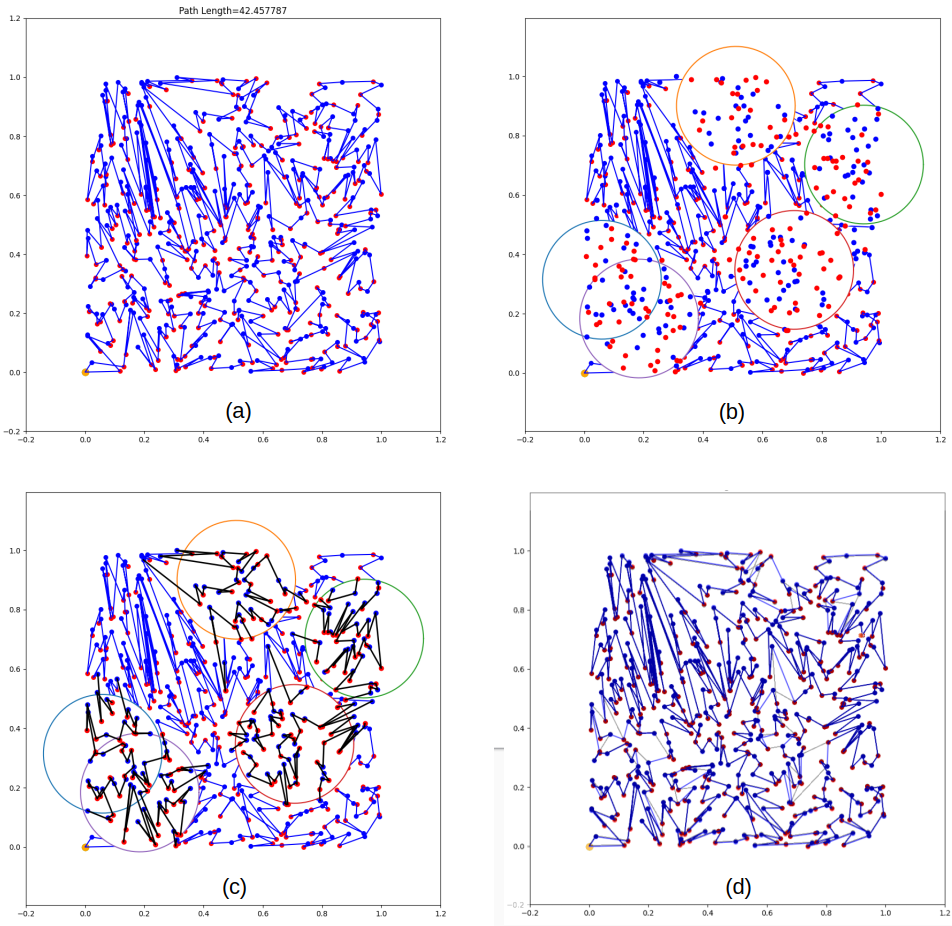}
\caption{Polishing Path Solution with SLPPR:Examples with Five Circles. (a)Before polishing. (b).Breaking connections around the circles. (c).Optimized reconnection with SLPPR. (d). Reconstructed tour with less distance cost.}
\label{fig:5c}
\end{figure}

This spatially adaptive strategy allows the algorithm to focus computational effort on local subregions of interest, providing a mechanism for incremental improvement and fine-scale correction of the global routing solution when the algorithm is applied iteratively along the tour path. Meanwhile, when the computation resource from the hardware is sufficient, such method can be extended to apply multiple domain nodes selections for path breaking and reconnection. The domain filter also can be general shape, and the number of domain of interests can also be more than one. For example, Figure \ref{fig:5c} presents an example of polishing the tour path with 5 circles, and final cycle is optimized by \textbf{Large-$\alpha$} method as introduced in Section 2.6. While multiple domain of interests involves more complex issues in: computational resource limitation; effective domain identification; effects on efficiency and accuracy of methods etc., which is not within the main scope of this paper and thus left for future work. This work study the property of algorithms with single circle based SLPPR for path polishing and further improvement.  

\subsubsection{Determination of the Number of Refinement Steps}

The step length of Local-PPR iterations, denoted by $n_{\mathrm{stp}}$, is determined according to the spatial point density and the refinement radius $r_r$.  
This parameter controls how many localized refinement operations are applied sequentially within the domain, effectively balancing coverage and computational cost.

\paragraph{Characteristic Node Spacing.}
Let the total number of nodes in the problem be $n$, approximately assume  that they are uniformly distributed over a planar region of area $S$.  
The characteristic inter-node spacing $d_s$ can be approximated by
\begin{equation}
d_s = \sqrt{\frac{S}{n}}.
\label{eq:spacing}
\end{equation}
This distance represents the average separation between adjacent nodes under uniform density assumptions.

\paragraph{Refinement Step Definition.}
For a refinement region of radius $r_r$, a single local reconstruction step effectively covers a neighborhood of approximately $2r_r$ in diameter.  
To ensure appropriate coverage of the entire domain, the total number of steps is thus proportional to the ratio between the characteristic length of a refinement region and the average node spacing:
\begin{equation}
n_{\mathrm{stp}} = 2 \, \kappa \, \frac{r_r}{d_s}
= 2 \, \kappa \, \frac{r_r}{\sqrt{S/n}}=\eta ,
\label{eq:nstp}
\end{equation}
where $r_r$ is defined in Equation \ref{eq:radius-simple} and $\kappa$ is a dimensionless proportionality coefficient that controls the degree of overlap or redundancy between successive refinement regions. Therefore, the number of refinement steps $n_{\mathrm{stp}}$ is a independent on problem size $n$.

The coefficient \texttt{$\eta$} is chosen  to balance overlap and computational efficiency. $n_{stp}=\eta=3$ is used in the experiments.
Beginning from staring points, the solver polish the solution along the path every $n_{\mathrm{stp}}$ steps, ensuring that the Local-PPR process systematically and efficiently explores the entire domain to improve the solution quality after proper merging solution is produced.

\subsection{Constraints for Near-Optimal Path}
After polishing the solution with Local-PPR along the entire path, the solution is already very near to optimal solution in terms of path length cost. However, there are still space for further improvements. Here, let's the $\alpha$ be the percentage of edges difference between current solution and the ground truth path. When $\alpha$ is very low, it meaning that current solution is having very high percentage of valid edges in global optimal solution (Global optimal solution could have multiple manner with the same cost, here the one that have least number of different edges makes the most sense for further improving current solution). 

Therefore, when a very high quality near-to-optimal-solution is achieved, the edges in the solution can be determined using Equation \ref{eq:initial-cycle-edges} and Equation \ref{eq:c2ip}. Let the complete edges set for the complete solution be $L_s$

\[
L_s = \{ (i,j) \in E : x_{ij} = 1 \}.
\]
Based on the fact that the solution is near-to-optimal, \textbf{the idea is to further improve the solution based on the existing near-optimal-solution constraints}. 

To ensure $(1- \alpha)$ percent of the edges are kept in resulting solution, following constraints are defined:

\begin{equation}
\sum_{(i,j) \in L_s} x_{ij} \geq 2n(1-\alpha).
\label{eq:edge-sum}
\end{equation}

Equation~\eqref{eq:edge-sum} guarantees that at least $int(2n(1-\alpha))$ of the original edges are maintained among the candidates in $L_s$ , while other connections can be updated to achieve better solution results. Such constraint is named as \textbf{Large-$\alpha$} constraint. 
Here, the JAR solver is used, with additional constraints introduced here to speed up the problem solving.

\subsubsection{Comparing with k-opt method} Such constraint is relevant to k-opt algorithm. As known that k-opt algorithm attempts to swap $k$ links in the tour to make possible improvements. In other words, $k$ edges out of the $n$ (for typical TSP, for JAR problem, it will be $2n$). Therefore, the constraint reflected from k-opt algorithm corresponding to the JAR problem is(considering there are items and placeholder types with more sophisticated constraints, such formulation is an approximation only):
\begin{equation}
\sum_{(i,j) \in L_s} x_{ij} = 2n-k.
\label{eq:edge-k}
\end{equation}

Comparing Equation \ref{eq:edge-sum} with \ref{eq:edge-k}, it is clear that the \textbf{Large-$\alpha$} constraint enables more path update opportunities towards more optimal solutions. Though k-opt is efficient when k is small number, the experimental results (see Table \ref{tab:path_results} and \ref{tab:path_results_500}) show that for large size problem (300,500 for example), the number of edge exchanges could be large (some times over hundred) in order to improve the path quality, while using k-opt with a smaller number will not lead to any progress in the path optimization. In other words, when a tour is k-optimal path, it is not necessarily global optimal. Therefore, improving the path involves larger number of edges to be exchanged. 
Such evidence indicates the limitation of k-opt methods.

\subsubsection{Parameter Sensitivity and Solver Efficiency}

With \textbf{Large-$\alpha$} constraints, it actually covers all the possible edge exchanges under $2n \times \alpha$. 
Experimental results for $n=300$ and $n=500$ with different $\alpha$ values are presented in Tables~\ref{tab:path_results} and~\ref{tab:path_results_500}. 

For large-scale \( K \)-opt problems (for instance, \( n = 1000 \) and $\alpha$ = 0.1 , the number of potential edge exchanges can reach approximately 200 over a total of about 2000 edges. Such instances become computationally and memory intensive to solve, especially when formulated as a mixed-integer optimization problem. The combinatorial explosion of possible edge changes results in a very large model size, making it challenging for solvers such as \textsc{Gurobi} to handle efficiently within practical time and memory limits. Therefore, for extremely large size problem, achieving \textbf{exact solution} with high efficiency is still challenging. In such cases, the introduced polishing method is useful in finding practically almost optimal solutions. 
Then, in path refinement with \textbf{Large-$\alpha$} constraints, the parameter $\alpha$  can be set to a value smaller than the actual deviation from the optimal solution in order to ensure that the problem is still within the capacity of the computer hardware, particularly for large-scale instances when obtaining an exact optimum is less critical. 

The impact of varying $\alpha$ on both solution quality and computational efficiency for large problem sizes remains an open area of research. In particular, when multiple optimization runs are involved, careful tuning of $\alpha$ may enhance the solver’s capability while increasing the methodological complexity. This trade-off between efficiency and accuracy represents a interesting direction for future investigation.

% Formally, the resulting PPR optimization model seeks to determine the optimal set of new edges
% \[
% L_{\mathrm{new}} = \arg\min_{x_{ij} \in \{0,1\}} 
%     \sum_{(i,j)\in E'} c_{ij} x_{ij},
% \]
% subject to:
% \[
% x_{ij} = 1, \quad \forall (i,j) \in L_r, 
% \qquad
% x_{ij} = 0, \quad \forall (i,j) \notin E',
% \]
% where $E'$ includes only edges between the boundary nodes and their feasible placeholder counterparts.

\begin{figure}[h!]
\centering    \includegraphics[width=0.6\textwidth]{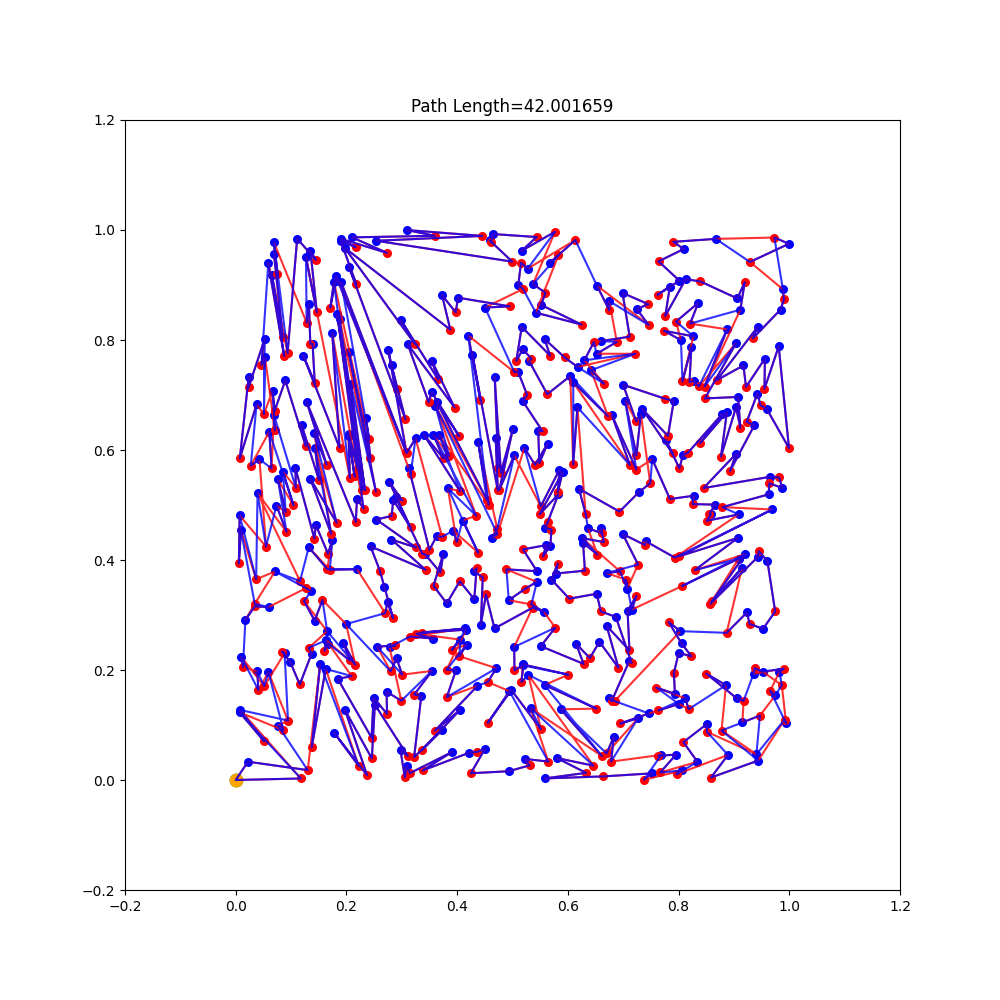}
\caption{Sample of Final Optimal Path for n=300 (in blue) (Merged Initial Guess is in Red).}
\label{fig:Merge2Final}
\end{figure}

\section{Computational Complexity}
In \cite{yuan2025datasets,basu2020complexitybranchandboundcuttingplanes}, it is mentioned that the JAR solver with linear programming and cutting-plane method is of exponential computation complexity. 
\paragraph{Two-Way Assignment} As discussed in \cite{qilong2025assignmentroutingoptimizationefficient}, the two-way assignment applies Hungarian algorithms to in two steps, resulting a complexity of $O(n^3)$, which is very quickly solved with negligible time (0.1(???) second for n=1000) as compared with other processes in the methods.
\paragraph{Initial Merging Complexity}

As discussed in Table\ref{tab:cycle_percentage_combined} that for JAR problem with size $n$, the resulting cycle from the two-way assignment method is $w_c \dot n$, $w_c <1$ is the cycle number to pair number ratio. It is obvious that the computational complexity for initial merging $O_I$, 
\begin{equation}
O_I<O((w_c \dot n)^2*n^2)<O(n^4)
\end{equation}

\paragraph{Merging path refinement with partial path reconstruction}

As introduced in Section 2.4, the PPR based merging path refinement is solving a JAR problem with much reduced size $w_p \dot n$. Since JAR is of exponential complexity, such PPR based merging refinement is still costly in computation when $n$ is large, as shown in Table \ref{tab:n300_avgpct} and \ref{tab:n500_avgpct}.

\paragraph{Spatially localized partial path reconstruction}
Knowing that the localized PPR is covering small size problem using JAR method, which can be solved in finite steps. Lets assume the maximum computation steps is $N_m$, given problem size $n$ and item step length  $n_{stp}$, the computational complexity is actually linear with respect to problem size:
\begin{equation}
O(N_m*N/n_{stp}) \rightarrow O(n)
\end{equation}

\paragraph{JAR with \textbf{Large-$\alpha$} constraints}
Though a rigorous analysis of the computational complexity of JAR with \textbf{Large-$\alpha$} constraints method (\textbf{Large-$\alpha$} method for short) is not presented, experiments results shows such method works efficiently for moderate size problems ($n=300$ and $n=500$ with $\alpha$=0.015,0.05,and 0.15 separately as shown in Table \ref{tab:polish_large} and \ref{tab:polish_large_n500}). For large size problem, $\alpha$ need to be set much smaller in order to complete the problem solving correctly. For ($n=1000$ , $\alpha$=0.015 works, while when $\alpha$ =0.05,and 0.15, the solver falls before completing the problem solving, \textbf{Even when the near-optimal solution is provided as start value}. 
From Appendix section, it is clear that if instead of applying $k=2n \times \alpha$, let $k$ be a constant. Then, it becomes a problem with polynomial computational  complexity $O(n^k)$. However, as Table \ref{tab:cycle_percentage_combined} reflects that such k can be very large number and thus the problem would not computational heavy when n is large. As a results only small $\alpha$ works for large $n$ problem. However, such solution from small $\alpha$ (for example: we would want to explore $\beta=\alpha/log(n)$ to replace $\alpha$ for testing) would not likely to be the optimal solution due to the increasing complexity of the problem when n increase.  Such facts underscores the actual challenge of solving JAR problem with larger sizes. 

% Reduces feasible region in every LP relaxation — the LP now starts closer to your initial tour.

% Tightens bounds, because many fractional solutions violating this constraint are removed.

% Improves branching efficiency, since a lot of “far away” subproblems are eliminated.
% Appendix section introduces the estimated complexity of such method referring to computational complexity for k-opt searching within a complete tour. 

\section{Experimental Results}
\paragraph{Computer Hardware}
All experiments were conducted on a workstation equipped with an Intel(R) Core(TM) Ultra 7 155H processor, supporting the SSE2, AVX, and AVX2 instruction sets. The CPU has 22 physical cores and 22 logical processors.

\paragraph{Datasets and Experiments} To study the performance of proposed methods:
Dataset from \cite{yuan2025datasets} with n=300, n=500, and n=1000 are applied in the experimental testing. For each size group, the first 10 samples are tested in the experiments. Gurobi ground truth is applied as the benchmark reference for $n=300$ and  $n=500$. For $n=1000$, no ground truth reference is currently available. The results from Heuristic method \cite{qilong2025assignmentroutingoptimizationefficient} are referred to for comparison purpose. 

% Following experiments are conducted. 
% First experiments apply solver with double-sided Hungarian assignment + fast cycle merging + JAR enhanced cycle merging + Local-PPR + Large-constraint based JAR. 

% Second experiments apply solver with double-sided Hungarian assignment + fast cycle merging +  Local-PPR . 

\subsection {Experiments on Merging Data Properties} 
First of all, to study the efficiency of merging algorithm, we investigate the number of cycles for each group, as shown in Table \ref{tab:cycle_percentage_combined}, which indicates that the number of cycles is around 20\% of the size of item-placeholder pairs. Meanwhile, the resulting releasing nodes from merging is around $1/3$ of total node size which is $2n+2$, including the start and stop nodes. Such data evidences show that applying JAR method for enhanced cycle merging is solving a problem with is only around $1/3$ of the problem size, and thus can be solved more efficiently.

\begin{table}[htbp]
\footnotesize
\centering
\caption{Cycle lengths and node points (with percentage of $2n+2$) for different sizes of problems}
\label{tab:cycle_percentage_combined}

\begin{tabular}{
    c |
    S[table-format=3.0] l |
    S[table-format=3.0] l |
    S[table-format=3.0] l
}
\toprule
\textbf{Index} & 
\multicolumn{2}{c|}{$n = 300$} &
\multicolumn{2}{c|}{$n = 500$} &
\multicolumn{2}{c}{$n = 1000$} \\
\cmidrule(lr){2-3} \cmidrule(lr){4-5} \cmidrule(lr){6-7}
 & \textbf{len\_cycles} & \textbf{node\_points} 
 & \textbf{len\_cycles} & \textbf{node\_points} 
 & \textbf{len\_cycles} & \textbf{node\_points} \\
\midrule
1  & 59 & 201 (33.3\%) & 99 & 347 (34.6\%) & 169 & 593 (29.6\%) \\
2  & 69 & 238 (39.6\%) & 96 & 331 (33.0\%) & 165 & 583 (29.1\%) \\
3  & 57 & 196 (32.6\%) & 102 & 343 (34.1\%) & 200 & 682 (34.1\%) \\
4  & 51 & 169 (28.1\%) & 82 & 297 (29.5\%) & 200 & 700 (35.0\%) \\
5  & 52 & 185 (30.8\%) & 77 & 261 (25.9\%) & 192 & 662 (33.1\%) \\
6  & 52 & 175 (29.2\%) & 84 & 287 (28.4\%) & 174 & 604 (30.2\%) \\
7  & 67 & 223 (37.2\%) & 96 & 340 (33.7\%) & 170 & 606 (30.3\%) \\
8  & 57 & 193 (32.2\%) & 77 & 271 (26.8\%) & 163 & 565 (28.3\%) \\
9  & 65 & 218 (36.4\%) & 79 & 271 (26.8\%) & 191 & 655 (32.8\%) \\
10 & 56 & 189 (31.5\%) & 94 & 324 (32.3\%) & 197 & 665 (33.2\%) \\
\midrule
\textbf{Average} & 
\multicolumn{1}{c}{58.7 (0.196)} & \multicolumn{1}{c|}{199.2 (33.1\%)} &
\multicolumn{1}{c}{90.3 (0.181)} & \multicolumn{1}{c|}{306.2 (30.4\%)} &
\multicolumn{1}{c}{181.1 (0.181)} & \multicolumn{1}{c}{631.5 (31.6\%)} \\
\bottomrule
\end{tabular}
\end{table}

\sisetup{
    round-mode=places,
    round-precision=4,
    table-number-alignment=center
}

\begin{table}[htbp]
\footnotesize
\centering
\caption{Results for $n = 300$: Comparison of Gurobi, Merge, and JAR-Merge Methods. Percentage difference from Gurobi is shown in parentheses.}
\label{tab:n300_avgpct}

\begin{tabular}{
    c |
    r r |
    l r |
    l r
}
\toprule
\textbf{Index} &
\multicolumn{2}{c|}{\textbf{Gurobi}} &
\multicolumn{2}{c|}{\textbf{Merge}} &
\multicolumn{2}{c}{\textbf{JAR-Merge}} \\
\cmidrule(lr){2-3} \cmidrule(lr){4-5} \cmidrule(lr){6-7}
 & L\_Gurobi & dt\_gurobi & L\_merge (\%$\delta$) & dt\_merge & L\_JAR\_merge (\%$\delta$) & dt\_JAR\_merge \\
\midrule
1 & 42.0017 & 76.0503 & 42.4578 (+1.086\%) & 2.2297 & 42.2880 (+0.682\%) & 11.9506 \\
2 & 33.1289 & 25.6637 & 33.7769 (+1.957\%) & 2.7524 & 33.3882 (+0.783\%) & 26.0197 \\
3 & 34.3303 & 44.0879 & 34.8066 (+1.386\%) & 1.9997 & 34.6356 (+0.887\%) & 7.8358 \\
4 & 35.1968 & 28.5358 & 35.6473 (+1.278\%) & 1.9571 & 35.5088 (+0.884\%) & 4.4184 \\
5 & 33.8118 & 77.8469 & 34.2057 (+1.166\%) & 1.8055 & 34.1478 (+0.997\%) & 4.4032 \\
6 & 37.4785 & 84.7612 & 37.8155 (+0.900\%) & 1.9396 & 37.6872 (+0.562\%) & 5.9695 \\
7 & 33.9530 & 26.5168 & 34.5650 (+1.803\%) & 1.9893 & 34.1743 (+0.653\%) & 7.3292 \\
8 & 40.6682 & 28.3229 & 40.9081 (+0.590\%) & 2.0084 & 40.7847 (+0.288\%) & 5.9320 \\
9 & 36.6766 & 51.3296 & 37.1410 (+1.266\%) & 2.1064 & 36.9246 (+0.672\%) & 8.2235 \\
10 & 38.6581 & 83.7323 & 38.9472 (+0.748\%) & 2.0598 & 38.8468 (+0.485\%) & 5.5707 \\
\midrule
\textbf{Average} & - & - & (+1.118\%) & - & (+0.689\%) & - \\
\bottomrule
\end{tabular}
\end{table}

\begin{table}[htbp]
\footnotesize
\centering
\caption{Results for $n = 500$: Comparison of Gurobi, Merge, and JAR-Merge Methods. Percentage difference from Gurobi is shown in parentheses.}
\label{tab:n500_avgpct}

\begin{tabular}{
    c |
    r r |
    l r |
    l r
}
\toprule
\textbf{Index} &
\multicolumn{2}{c|}{\textbf{Gurobi}} &
\multicolumn{2}{c|}{\textbf{Merge}} &
\multicolumn{2}{c}{\textbf{JAR-Merge}} \\
\cmidrule(lr){2-3} \cmidrule(lr){4-5} \cmidrule(lr){6-7}
 & L\_Gurobi & dt\_gurobi & L\_merge (\%$\delta$) & dt\_merge & L\_JAR\_merge (\%$\delta$) & dt\_JAR\_merge \\
\midrule
1 & 41.4737 & 178.9940 & 42.0443 (+1.377\%) & 8.4204 & 41.8935 (+1.004\%) & 71.0571 \\
2 & 47.4789 & 144.3976 & 47.9567 (+1.011\%) & 8.8708 & 47.8406 (+0.789\%) & 36.2083 \\
3 & 49.6166 & 120.3259 & 50.0434 (+0.855\%) & 7.5352 & 49.8723 (+0.508\%) & 93.6267 \\
4 & 57.7156 & 273.2535 & 58.1085 (+0.673\%) & 6.5373 & 58.0152 (+0.518\%) & 18.9444 \\
5 & 55.0588 & 85.2649 & 55.3181 (+0.474\%) & 8.5678 & 55.2404 (+0.335\%) & 11.2766 \\
6 & 56.2540 & 160.3211 & 56.6719 (+0.737\%) & 8.3486 & 56.5319 (+0.492\%) & 13.3917 \\
7 & 55.0596 & 155.0151 & 55.4291 (+0.669\%) & 9.4675 & 55.3184 (+0.471\%) & 33.7567 \\
8 & 51.3383 & 130.7341 & 51.7296 (+0.760\%) & 7.7906 & 51.5505 (+0.419\%) & 9.1243 \\
9 & 43.2535 & 17.7952 & 43.6961 (+1.017\%) & 7.5601 & 43.5802 (+0.756\%) & 26.1633 \\
10 & 42.6967 & 124.6362 & 43.3448 (+1.500\%) & 7.4173 & 43.1552 (+1.070\%) & 36.8638 \\
\midrule
\textbf{Average} & - & - & (+0.978\%) & - & (+0.633\%) & - \\
\bottomrule
\end{tabular}
\end{table}

Table \ref{tab:n300_avgpct} and \ref{tab:n500_avgpct} shows the results of cycle merging for $n=300$ and $n=500$ separately, with $L\_{gurobi}$,$L\_{merge}$, and $L\_{JAR}\_{merge}$
indicating the tour cost from the gurobi ground truth, the fast initial merging algorithm as introduced in Section 2.3 and the merging optimization based on PPR method as introduced in Section 2.4.  Results show that the accuracy of merging path increase from 1\% to 0.6\% after the merging optimization is applied, for  $n=300$ , $n=500$ ( $n=1000$ also show matching results as shown in Table \ref{tab:n1000_new_heuristic_merge}). It is also clearly shown that the JAR merging time is much faster as compared with the complete problem solving time in Gurobi solver. 
\subsection{Experiments to Study Performance of SLPPR and \textbf{Large}-$\alpha$ Methods}
Such JAR based merging enhancement solver provides very high quality solutions with slightly less deviation error as compared with the results from \cite{qilong2025assignmentroutingoptimizationefficient}. However, the speed is much slower. Especially when $n$ is large, such size-reduced  JAR problem solving would still be computationally heavy. Therefore, we design the following comparison experiments to study whether JAR based merging step can be replaced by more efficient SLPPR algorithm to save computation time.
\begin{enumerate}
    \item Set A: Solve problem with Initial fast merging, followed by JAR-Merging, and then followed by SLPPR. 
    \item Set B: Solve problem with Initial fast merging,  followed directly by SLPPR.
\end{enumerate}

It is noticed that after the SLPPR algorithms is applied, the solution will be the same no matter whether the JAR-Merging algorithm is included or not. This is due to the fact that SLPPR actually release the local connections for better connection with more nodes releasing and reconnection. Since SLPPR is applied along the entire tour cycle, such path improvement is much stronger than the JAR-Merging approach.

Table \ref{tab:polish_large} and \ref{tab:polish_large_n500} shows the experimental results with serial of solvers as follows:

% Initial merging \rightarrow SLPPR polishing solver, 1st and 2nd. 
%     \rightarrow{Large-\alpha} solver

% --- Vertically and horizontally centered flowchart ---
\thispagestyle{empty} % no page number
\begin{center}
\begin{tikzpicture}[
    node distance = 1.0cm,
    >=Stealth,
    every node/.style={font=\footnotesize\sffamily},
    process/.style={
        rectangle,
        rounded corners,
        draw=black,
        thick,
        minimum width=3.2cm,
        minimum height=1.2cm,
        text centered,
        align=center,
        fill=gray!10
    },
    arrow/.style={thick, ->, >=stealth}
]

% --- Nodes ---
\node[process] (merge) {Initial\\Merging};
\node[process, right=of merge] (slppr1) {1st SLPPR\\Polishing Solver\\};-\
\node[process, right=of slppr1] (slppr2) {2nd SLPPR\\Polishing Solver\\};
\node[process, right=of slppr2] (large) {Large-$\alpha$\\Solver};

% --- Arrows ---
\draw[arrow] (merge) -- (slppr1);
\draw[arrow] (slppr1) -- (slppr2);
\draw[arrow] (slppr2) -- (large);

\end{tikzpicture}
\end{center}

Initial guess solvers is directly connected to a SLPPR polishing solver, followed by 2 second SLPPA solver and a Large-$\alpha$ solver. After the first SLPPR polishing solver, the deviation error is reduced to \textbf{0.09\% } for both $n=300$ and $n=500$ sets. Therefore, it is more powerful in achieving higher quality solutions within shorter time. 

In the experiments, the polishing radius $r_r$ is set to be 0.2m for the experiments. A second polishing still slightly improves the solution with error deviations reduced to 0.07\%. While we have tested that more then 2 times polishing would not help in improving the solution in the experiment datasets. 

After SLPPR polishing, the results are further processed by \textbf{Large-$\alpha$} solvers, the progress efficiency and performance is much dependent on the selection of $\alpha$ value. As shown in Table \ref{tab:path_results} and \ref{tab:path_results_500}, with the increase of $\alpha$ value from 0.015 to 0.05 and then to 0.15, the deviation reduced from 0.07\% to 0.00\%, while the average number of edge differences from ground truth references are also much reduced.

\begin{table}[htbp]
\footnotesize
\centering
\caption{Comparison of Gurobi, Polish, and Large methods. Percentage difference from Gurobi is shown in parentheses.}
\label{tab:polish_large}

\begin{tabular}{
    c |
    r |
    l r |
    l r |
    l r
}
\toprule
\textbf{Index} &
\textbf{L\_Gurobi} &
\multicolumn{2}{c|}{\textbf{Polish1}} &
\multicolumn{2}{c|}{\textbf{Polish2}} &
\multicolumn{2}{c}{\textbf{Large}} \\
\cmidrule(lr){3-4} \cmidrule(lr){5-6} \cmidrule(lr){7-8}
 &  & L\_Polish1 (\%$\delta$) & t\_polish1 & L\_Polish2 (\%$\delta$) & t\_polish2 & L\_large (\%$\delta$) & t\_L \\
\midrule
1 & 42.0017 & 42.0131 (+0.027\%) & 7.1489 & 42.0116 (+0.024\%) & 6.8130 & 42.0116 (+0.024\%) & 17.0288 \\
2 & 33.1289 & 33.2001 (+0.214\%) & 6.7550 & 33.1873 (+0.176\%) & 5.9664 & 33.1789 (+0.151\%) & 16.5580 \\
3 & 34.3303 & 34.4001 (+0.203\%) & 6.3549 & 34.3476 (+0.050\%) & 5.9866 & 34.3378 (+0.022\%) & 14.4317 \\
4 & 35.1968 & 35.2156 (+0.053\%) & 6.0150 & 35.2156 (+0.053\%) & 5.8511 & 35.2156 (+0.053\%) & 19.2373 \\
5 & 33.8118 & 33.8693 (+0.170\%) & 6.7989 & 33.8656 (+0.160\%) & 6.2295 & 33.8656 (+0.160\%) & 22.3974 \\
6 & 37.4785 & 37.4785 (+0.000\%) & 5.9428 & 37.4785 (+0.000\%) & 5.8332 & 37.4785 (+0.000\%) & 17.3717 \\
7 & 33.9530 & 34.0142 (+0.180\%) & 6.6118 & 34.0142 (+0.180\%) & 6.8551 & 34.0142 (+0.180\%) & 19.7932 \\
8 & 40.6682 & 40.6921 (+0.059\%) & 6.2423 & 40.6926 (+0.060\%) & 5.7592 & 40.6834 (+0.037\%) & 12.0455 \\
9 & 36.6766 & 36.6857 (+0.025\%) & 6.8033 & 36.6766 (+0.000\%) & 6.5106 & 36.6766 (+0.000\%) & 16.1633 \\
10 & 38.6581 & 38.6609 (+0.007\%) & 6.5016 & 38.6609 (+0.008\%) & 6.3440 & 38.6586 (+0.001\%) & 22.9639 \\
\midrule
\textbf{Average} & - & (+0.094\%) & - & (+0.071\%) & - & (+0.063\%) & - \\
\bottomrule
\end{tabular}
\end{table}

\begin{table}[htbp]
\footnotesize
\centering
\caption{Results for $n = 500$: Comparison of Gurobi, Polish, and Large methods. Percentage difference from Gurobi is shown in parentheses.}
\label{tab:polish_large_n500}

\begin{tabular}{
    c |
    r |
    l r |
    l r |
    l r
}
\toprule
\textbf{Index} &
\textbf{L\_Gurobi} &
\multicolumn{2}{c|}{\textbf{Polish1}} &
\multicolumn{2}{c|}{\textbf{Polish2}} &
\multicolumn{2}{c}{\textbf{Large}} \\
\cmidrule(lr){3-4} \cmidrule(lr){5-6} \cmidrule(lr){7-8}
 &  & L\_Polish1 (\%$\delta$) & t\_polish1 & L\_Polish2 (\%$\delta$) & t\_polish2 & L\_large (\%$\delta$) & t\_L \\
\midrule
1  & 41.4737 & 41.5559 (+0.198\%) & 11.0976 & 41.5334 (+0.144\%) & 10.3516 & 41.5209 (+0.114\%) & 60.4060 \\
2  & 47.4789 & 47.5234 (+0.094\%) & 10.9138 & 47.4993 (+0.043\%) & 9.6561  & 47.4932 (+0.030\%) & 40.9263 \\
3  & 49.6166 & 49.6687 (+0.105\%) & 11.7511 & 49.6542 (+0.076\%) & 10.7542 & 49.6411 (+0.050\%) & 64.8112 \\
4  & 57.7156 & 57.7390 (+0.041\%) & 14.0585 & 57.7386 (+0.041\%) & 13.7853 & 57.7360 (+0.040\%) & 69.5994 \\
5  & 55.0588 & 55.0725 (+0.025\%) & 12.2057 & 55.0598 (+0.002\%) & 11.3861 & 55.0598 (+0.002\%) & 34.6338 \\
6  & 56.2540 & 56.3014 (+0.084\%) & 12.7654 & 56.2936 (+0.070\%) & 12.6022 & 56.2835 (+0.053\%) & 55.4543 \\
7  & 55.0596 & 55.1187 (+0.108\%) & 12.4410 & 55.1187 (+0.107\%) & 12.0854 & 55.1058 (+0.084\%) & 43.8032 \\
8  & 51.3383 & 51.3624 (+0.047\%) & 11.2642 & 51.3573 (+0.037\%) & 10.8014 & 51.3508 (+0.024\%) & 44.7896 \\
9  & 43.2535 & 43.2934 (+0.092\%) & 11.0939 & 43.2881 (+0.080\%) & 10.5642 & 43.2880 (+0.079\%) & 51.1875 \\
10 & 42.6967 & 42.7487 (+0.122\%) & 12.1031 & 42.7405 (+0.103\%) & 10.6200 & 42.7218 (+0.059\%) & 64.0795 \\
\midrule
\textbf{Average} & - & (+0.092\%) & - & (+0.070\%) & - & (+0.054\%) & - \\
\bottomrule
\end{tabular}
\end{table}

For n=1000, JAR solver fail to provide solutions for samples in the dataset and thus the dataset did not provide ground truth data for n=1000 and above as for now. However, Table \ref{tab:n1000_full} and \ref{tab:n1000_new_heuristic_merge} prove that the results from the proposed solutions outperform the Heuristic solution \cite{qilong2025assignmentroutingoptimizationefficient} by 0.6\%, and outperform the fast merging solution by 0.9\%, which indicates that the solutions would be much nearer to optimal ones.

\begin{table}[htbp]
\footnotesize
\centering
\caption{Results for $n = 1000$: Comparison of New, Heuristic, and Merge Methods. Percentage difference relative to L\_New is shown in parentheses.}
\label{tab:n1000_new_heuristic_merge}

\begin{tabular}{
    c |
    r |
    l |
    l
}
\toprule
\textbf{Index} &
\textbf{L\_Large} &
\textbf{L\_Heuristic (\%$\delta$)} &
\textbf{L\_Merge (\%$\delta$)} \\
\midrule
1  & 73.2240 & 73.6374 (+0.565\%) & 73.8331 (+0.833\%) \\
2  & 64.0881 & 64.4509 (+0.566\%) & 64.6675 (+0.904\%) \\
3  & 79.9331 & 80.3843 (+0.565\%) & 80.5530 (+0.777\%) \\
4  & 62.7093 & 63.2650 (+0.888\%) & 63.5323 (+1.313\%) \\
5  & 72.0800 & 72.4962 (+0.577\%) & 72.6829 (+0.837\%) \\
6  & 69.8928 & 70.3990 (+0.723\%) & 70.5887 (+0.996\%) \\
7  & 74.8898 & 75.2270 (+0.451\%) & 75.4329 (+0.726\%) \\
8  & 64.8747 & 65.2480 (+0.576\%) & 65.4273 (+0.852\%) \\
9  & 65.8594 & 66.3623 (+0.763\%) & 66.6030 (+1.129\%) \\
10 & 66.7659 & 67.1344 (+0.553\%) & 67.3441 (+0.865\%) \\
\midrule
\textbf{Average} & - & (+0.623\%) & (+0.933\%) \\
\bottomrule
\end{tabular}
\end{table}

\begin{table}[htbp]
\footnotesize
\centering
\caption{Comparison of $L_{\text{polish1}}$ and $L_{\text{polish2}}$ with respect to $L_{\text{Large}}$ for $n=1000$. 
Percentage difference from $L_{\text{Large}}$ shown in parentheses.}
\label{tab:n1000_full}

\begin{tabular}{c
S[table-format=3.4]
S[table-format=3.4]
l
S[table-format=3.4]
l
S[table-format=3.4]
S[table-format=3.4]}
\toprule
\textbf{Idx} & \textbf{$t_{\text{merge}}$} & \textbf{$t_{\text{polish1}}$} & \textbf{$L_{\text{polish1}}$ ($\delta$\%)} &
\textbf{$t_{\text{polish2}}$} & \textbf{$L_{\text{polish2}}$ ($\delta$\%)} &
\textbf{$L_{\text{Large}}$} & \textbf{$t_{\text{Large}}$} \\
\midrule
1  & 55.7264 & 25.4942 & 73.2537 (+0.040\%) & 24.1469 & 73.2325 (+0.011\%) & 73.2240 & 249.2742 \\
2  & 50.1367 & 26.7380 & 64.1198 (+0.050\%) & 25.8131 & 64.1198 (+0.050\%) & 64.0881 & 291.4471 \\
3  & 50.4052 & 27.3725 & 80.0160 (+0.103\%) & 24.9240 & 79.9488 (+0.020\%) & 79.9331 & 221.2972 \\
4  & 66.1459 & 25.5908 & 62.7336 (+0.039\%) & 23.5247 & 62.7157 (+0.010\%) & 62.7093 & 402.0683 \\
5  & 59.5681 & 27.6269 & 72.1202 (+0.055\%) & 24.3120 & 72.0948 (+0.020\%) & 72.0800 & 261.1203 \\
6  & 53.5351 & 25.0923 & 69.9398 (+0.067\%) & 23.2981 & 69.9288 (+0.052\%) & 69.8928 & 238.9883 \\
7  & 58.1689 & 26.2854 & 74.9454 (+0.074\%) & 24.8901 & 74.9104 (+0.028\%) & 74.8898 & 218.8815 \\
8  & 47.3178 & 24.9000 & 64.9286 (+0.083\%) & 22.7256 & 64.9030 (+0.044\%) & 64.8747 & 422.9316 \\
9  & 50.1000 & 28.8648 & 65.9218 (+0.095\%) & 25.1448 & 65.8731 (+0.036\%) & 65.8594 & 559.0728 \\
10 & 74.7135 & 24.8876 & 66.8083 (+0.064\%) & 23.2621 & 66.7878 (+0.033\%) & 66.7659 & 401.6271 \\
\midrule
\textbf{Average $\delta$} &
\multicolumn{1}{c}{ } &
\multicolumn{1}{c}{ } &
(+0.067\%) &
\multicolumn{1}{c}{ } &
(+0.030\%) &
\multicolumn{1}{c}{ } &
\multicolumn{1}{c}{ } \\
\bottomrule
\end{tabular}
\end{table}

Table \ref{tab:path_results} and \ref{tab:path_results_500} shows evidence that the number of edge connection difference  between near-to-optimal path to exact optimal can be large. Therefore, using k-opt algorithm to improve a near-optimal-solution to exact solution is very challenging when the problem size is large (when $n \geq 1000$). As noticed that solving such optimization problem can involve very large size of edge connection updates (could be over 100 for n=300 and n=500 problems), in which case k-opt methods would not be able to efficiently solve the problem. In other words, the k-optimal path (for example 30-opt, 50-opt, which is almost impossible to be applicable deal to super high computational complexity as shown in Appendix) is still having many different edges in the path. 
Based on experimental results, when the number of edges $k_{np}= 2n \times \alpha$  is large, the JAR problem solving is not much significantly less than solving the complete size $n$ JAR problem . Therefore, when an exact solution is not needed, and the efficiency is more critical, user should set a low $k_{np}$ in association with $\alpha$ value to make the solver work efficiently for a practically almost optimal solution (0.05\% deviation from exact solution, as indicated by the experimental results for the introduced methods). 

\textbf{Remarks: }
\textit{It is noted that the Gurobi have noticeable truncation errors in solving the problem with investigated experiment sizes. Based on complete edges comparison, path cost value can have noticeable difference around several times of 0.0001m, when the actual path is already exact solution.}

\begin{table}[H]
\footnotesize
\centering
\caption{Comparison of path length and runtime for different $\alpha$ values ($n=300$). Percentage differences are relative to $L_{\text{Gurobi}}$.}
\setlength{\tabcolsep}{4pt}
\renewcommand{\arraystretch}{1.2}
\begin{tabular}{c|c|ccr|ccr|ccr}
\hline
\multirow{2}{*}{Index} & \multirow{2}{*}{$L_{\text{Gurobi}}$} &
\multicolumn{3}{c|}{$\alpha = 0.015$} &
\multicolumn{3}{c|}{$\alpha = 0.05$} &
\multicolumn{3}{c}{$\alpha = 0.15$} \\
\cline{3-11}
 & & $L_{\text{large}}$ (\%$\Delta$) & $t_L$ & $N_d$ & $L_{\text{large}}$ (\%$\Delta$) & $t_L$ & $N_d$ & $L_{\text{large}}$ (\%$\Delta$) & $t_L$ & $N_d$ \\
\hline
0 & 42.0017 & 42.0116 (0.02\%) & 17.03 & 174 & 42.0095 (0.02\%) & 20.34 & 152 & 42.0024 (0.00\%) & 73.52 & 36 \\
1 & 33.1289 & 33.1789 (0.15\%) & 16.56 & 132 & 33.1576 (0.09\%) & 17.95 & 132 & 33.1289 (0.00\%) & 34.99 & 0 \\
2 & 34.3303 & 34.3378 (0.02\%) & 14.43 & 66 & 34.3307 (0.00\%) & 33.06 & 4 & 34.3309 (0.00\%) & 49.37 & 28 \\
3 & 35.1968 & 35.2156 (0.05\%) & 19.24 & 108 & 35.2035 (0.02\%) & 25.95 & 64 & 35.1968 (0.00\%) & 35.14 & 0 \\
4 & 33.8118 & 33.8656 (0.16\%) & 22.40 & 112 & 33.8301 (0.05\%) & 24.05 & 86 & 33.8118 (0.00\%) & 52.78 & 0 \\
5 & 37.4785 & 37.4785 (0.00\%) & 17.37 & 0 & 37.4785 (0.00\%) & 16.76 & 0 & 37.4785 (0.00\%) & 41.79 & 0 \\
6 & 33.9530 & 34.0142 (0.18\%) & 19.79 & 100 & 33.9723 (0.06\%) & 20.34 & 78 & 33.9530 (0.00\%) & 28.61 & 0 \\
7 & 40.6682 & 40.6834 (0.04\%) & 12.05 & 118 & 40.6715 (0.01\%) & 13.64 & 82 & 40.6684 (0.00\%) & 36.17 & 10 \\
8 & 36.6766 & 36.6766 (0.00\%) & 16.16 & 0 & 36.6766 (0.00\%) & 36.36 & 0 & 36.6766 (0.00\%) & 43.62 & 0 \\
9 & 38.6581 & 38.6586 (0.00\%) & 22.96 & 20 & 38.6581 (0.00\%) & 56.73 & 0 & 38.6581 (0.00\%) & 71.99 & 0 \\
\hline
\textbf{Avg.} & -- & -- (0.07\%) & 17.39 & 83 & -- (0.03\%) & 26.27 & 67 & -- (0.00\%) & 46.39 & 7 \\
\hline
\end{tabular}
\label{tab:path_results}
\end{table}

\begin{table}[H]
\footnotesize
\centering
\caption{Comparison of path length and runtime for different $\alpha$ values ($n=500$). Percentage differences are relative to $L_{\text{Gurobi}}$.}
\setlength{\tabcolsep}{4pt}
\renewcommand{\arraystretch}{1.2}
\begin{tabular}{c|c|ccr|ccr|ccr}
\hline
\multirow{2}{*}{Index} & \multirow{2}{*}{$L_{\text{Gurobi}}$} &
\multicolumn{3}{c|}{$\alpha = 0.015$} &
\multicolumn{3}{c|}{$\alpha = 0.05$} &
\multicolumn{3}{c}{$\alpha = 0.15$} \\
\cline{3-11}
 & & $L_{\text{large}}$ (\%$\Delta$) & $t_L$ & $N_d$ & $L_{\text{large}}$ (\%$\Delta$) & $t_L$ & $N_d$ & $L_{\text{large}}$ (\%$\Delta$) & $t_L$ & $N_d$ \\
\hline
0 & 41.4737 & 41.5209 (0.11\%) & 60.41 & 216 & 41.5034 (0.07\%) & 152.04 & 218 & 41.4737 (0.00\%) & 180.55 & 0 \\
1 & 47.4789 & 47.4932 (0.03\%) & 40.93 & 138 & 47.4798 (0.00\%) & 47.08 & 92 & 47.4788 (0.00\%) & 109.49 & 8 \\
2 & 49.6166 & 49.6411 (0.05\%) & 64.81 & 220 & 49.6321 (0.03\%) & 162.39 & 212 & 49.6184 (0.00\%) & 140.29 & 52 \\
3 & 57.7156 & 57.7360 (0.04\%) & 69.60 & 110 & 57.7214 (0.01\%) & 43.59 & 44 & 57.7117 (0.00\%) & 160.79 & 142 \\
4 & 55.0588 & 55.0598 (0.00\%) & 34.63 & 176 & 55.0582 (0.00\%) & 35.03 & 168 & 55.0589 (0.00\%) & 88.29 & 176 \\
5 & 56.2540 & 56.2835 (0.05\%) & 55.45 & 174 & 56.2712 (0.03\%) & 160.34 & 94 & 56.2540 (0.00\%) & 136.69 & 0 \\
6 & 55.0596 & 55.1058 (0.08\%) & 43.80 & 270 & 55.0835 (0.04\%) & 152.75 & 272 & 55.0616 (0.00\%) & 178.04 & 48 \\
7 & 51.3383 & 51.3508 (0.02\%) & 44.79 & 252 & 51.3458 (0.01\%) & 139.54 & 206 & 51.3379 (0.00\%) & 120.96 & 96 \\
8 & 43.2535 & 43.2880 (0.08\%) & 51.19 & 108 & 43.2573 (0.01\%) & 42.02 & 22 & 43.2535 (0.00\%) & 37.25 & 0 \\
9 & 42.6967 & 42.7218 (0.06\%) & 64.08 & 114 & 42.6971 (0.00\%) & 43.92 & 8 & 42.6967 (0.00\%) & 89.13 & 0 \\
\hline
\textbf{Avg.} & -- & -- (0.05\%) & 52.87 & 178 & -- (0.02\%) & 97.48 & 134 & -- (0.00\%) & 124.15 & 52 \\
\hline
\end{tabular}
\label{tab:path_results_500}
\end{table}

\section{Conclusion}
\label{sec:conclusion}

This work presents an \textbf{almost optimal and highly efficient solver} for the Joint Routing--Assignment (JRA) problem. Starting from an initial guess generated via a fast merging-based approach, the author proposed \textbf{Partial Pair Reconnection (PPR)} method to effectively selects key item--placeholder with in domains of interest, to break edges within/around the domain and reconnect nodes with a optimizer in a larger scale than normal k-opt does, thereby substantially improving the quality of the routing path obviously. This targeted optimization greatly enhances the merging quality compared with previous merging solvers.

Furthermore, a \textbf{Sequential Local PPR (SLPPR)} strategy is introduced to iteratively polishes the complete tour along the path, progressively refining the solution toward near-optimal accuracy. Experimental evaluations conducted on instances with $n = 300$, $500$, and $1000$ demonstrate that a single round of polishing yields an average deviation (from exact solution) of only \textbf{0.09\%}, while two rounds SLPPR polishing reduce the deviation to \textbf{0.07\%}, achieving a near-optimal solution with low computational overhead.

In addition, the authors use experimental evidences to show that k-optimal path could be far from actual exact optimal solution, and thus, k-opt has large limitation in providing almost exact solution because large problem would need to change tens of and even hundreds of edges (for n=300, 500 examples) all together to effective progress towards optimal. Therefore, the author proposed a \textbf{Large-$\alpha$ solver} which is adding a Large-$\alpha$ constraint to improve the efficiency of JAR solver based on the fact that an achieved high quality path (<0.1\% deviation from exact solution, for example) would share same edges at high percentage. Adding such Large-$\alpha$ constraints
efficiently resolves the JRA problem obviously improved tour quality and reduced differences in edge connections from ground truth. T

Beyond the JRA problem, the proposed framework and methodologies exhibit strong potential for broader applications. The PPR and SLPPR mechanisms can be naturally extended to the classical Traveling Salesman Problem (TSP) and related routing and logistics optimization problems, including vehicle routing, robotic path planning, warehouse task assignment, and network design. The efficiency and scalability of the approach make it particularly suitable for large-scale combinatorial optimization tasks.

Future research will focus on extending the \textbf{Large-$\alpha$ solver} for multiple running trials to handle even larger problem sizes. Meanwhile, methods to identify multiple domains of interests and apply PPR based polishing methods efficiently to improve large scale problem solving will be investigated. 
Additionally, investigations on JAR problem under item type constraints and time-window constraints will be interesting topics to be studied.

\section*{Appendix: Computational Complexity of $k$-Opt Algorithm and  \textbf{Large-$\alpha$} Method}
Though k-opt is dealing with TSP with single type of nodes, it still make sense to understand its computational complexity and how it reflect the complexity of JAR method under \textbf{Large-$\alpha$} constraints. 

Let \( MT(k) \) denote the number of distinct \( k \)-opt move types. In a \( k \)-opt operation, removing \( k \) edges from a tour divides it into \( k \) path segments. To form a new valid tour, these \( k \) segments can be reconnected in different orders and orientations (some may be reversed). The total number of possible move types is determined by the number of inversions and permutations of the resulting segments.

Refering to \cite{Helsgaun2009GeneralKS},  can be computed as the product of the number of inversions of \( k-1 \) segments and the number of permutations of these segments:
\[
MT(k) = 2^{k-1} \times (k-1)!.
\]
Here, \( 2^{k-1} \) corresponds to the possible segment inversions (each of the \( k-1 \) segments can be reversed or not), and \( (k-1)! \) represents all possible permutations of the segments. 

The number of possible \( k \)-opt move types grows rapidly with \( k \). For example:
\[
MT(2) = 2, \quad MT(3) = 8, \quad MT(4) = 48, \quad MT(5) = 384.
\]
This combinatorial explosion illustrates why \( k \)-opt search is typically restricted to small \( k \) values in practical applications.
For finite $k$, \( MT(k) \) is constant value. However, when $MT(10) = 1.9e8$, $MT(50) = 3.4e77$, which is beyond the grab of computational power. 
Moreover, for a problem with \( n \) nodes, the number of ways to choose \( K \) edges from the current tour is:
$\binom{n}{K}$.

 Thus, the total size of the neighborhood explored by a full K-opt move is roughly:
\[
O\left(\binom{n}{k} \times 2^(k-1) \times (k - 1)!\right)=O\left(\binom{n}{k} \times MT(k)\right),
\]
which grows with both $n$ and $k$.

\begin{table}[h!]
\centering
\begin{tabular}{@{}lll@{}}
\toprule
\textbf{K value} & \textbf{Approximate Complexity} & \textbf{Remarks} \\
\midrule
2 & \( O(n^2) \) & Efficient and commonly used (2-opt) \\
3 & \( O(n^3) \) & Still feasible; improves quality (3-opt) \\
4 & \( O(n^4) \) & Computationally heavy \\
5+ & \( O(n^K) \) or higher & Infeasible for large \( n \) \\
\bottomrule
\end{tabular}
\caption{Growth in computational complexity with increasing K in K-opt algorithms.}
\label{tab:kopt_complexity}
\end{table}

In practice, exploring the full K-opt neighborhood is rarely feasible for \( k > 3 \), especially when \( n \) is large. To maintain tractability, heuristic or restricted variants such as the Lin–Kernighan heuristic are typically employed. These methods limit the number of candidate edges or apply adaptive neighborhood selection to achieve near-optimal solutions with polynomial-time complexity.

\paragraph{Complexity of Large-$\alpha$ Constraints}
In the proposed framework, the parameter $\alpha$  controls the extent of edge exchanges allowed during optimization. When Large-$\alpha$ constraints are applied, all possible $k$-opt moves up to
\[
k_{\max} = 2n\alpha
\]
are considered, meaning that every possible exchange involving up to $k_{\max}$ edges is included in the model.

For each $k$, there are $\binom{2n}{k}$ ways to choose the edges to remove and $MT(k)$ possible reconnection types. Thus, the total number of potential exchange configurations is:
\[
N_{\text{total}}(\alpha, n) = \sum_{k=2}^{2n\alpha} \binom{2n}{k} \, MT(k).
\]

Using the approximation
\[
\binom{2n}{k} \approx \frac{(2n)^k}{k!} \quad \text{for } k \ll 2n,
\]
we obtain
\[
N_{\text{total}}(\alpha, n) 
\approx \sum_{k=2}^{2n\alpha} \frac{(2n)^k}{k!} \, 2^{k-1}(k-1)! 
= \frac{1}{2} \sum_{k=2}^{2n\alpha} (4n)^k \frac{1}{k}.
\]
The asymptotic order of growth is therefore:
\[
N_{\text{total}}(\alpha, n) = O\!\left((4n)^{2n\alpha}\right),
\]
indicating super-exponential complexity with respect to both $n$ and $\alpha$, which is in the same order of complexity with complete k-opt searching.

As $n$ and $\alpha$ increases, the number of potential exchange configurations grows explosively. For moderate size of problem with $n \in [300,500]$, the JAR problem is quite solvable, and setting a proper $\alpha$ is useful to push the solution to optimal with computational time comparable shorter than directly solving JAR without constraints.

For large size problem,$\alpha$ need to be set very small in order to ensure that the solver can handle the problem solving without breaking. 
Currently, solving the JAR problem for size $n=1000$ with \textbf{Large-$\alpha$} method is a challenge, which is an open problem remains to be resolved. 

\paragraph{Use of Generative AI.}
This manuscript involved the use of generative AI tools (ChatGPT-4, OpenAI) during the drafting process. These tools were employed to assist with language refinement, equation formatting, and improving the clarity of technical explanations. All scientific content, methods, experimental results, and conclusions were developed and validated by the authors. Final manuscript versions were thoroughly reviewed to ensure accuracy and originality.

\bibliographystyle{unsrt}  % or another style
%\bibliography{references}  % assuming references.bib

\end{document}